\pdfoutput=1
\documentclass[twoside]{article}
\usepackage[accepted]{aistats2018}

\usepackage[utf8]{inputenc} 
\usepackage[T1]{fontenc}    
\usepackage{hyperref}       
\usepackage{url}            
\usepackage{booktabs}       
\usepackage{amsfonts}       
\usepackage{nicefrac}       
\usepackage{microtype}      
\usepackage{amsmath}
\usepackage{graphicx}
\usepackage{subfig}
\usepackage[round]{natbib}

\newcommand{\bigO}{\mathcal{O}}
\newcommand{\R}{\mathbb{R}}
\newcommand{\N}{\mathcal{N}}
\newcommand{\E}{\mathbb{E}}
\newcommand{\Knn}{K_{nn}}
\newcommand{\Knm}{K_{nm}}
\newcommand{\Kmn}{K_{mn}}
\newcommand{\Kmm}{K_{mm}}
\newcommand{\ki}{k_{i}}
\newcommand{\KL}{\mbox{KL}}
\newcommand{\tr}{\mbox{tr}}

\begin{document}
\twocolumn[

\aistatstitle{Scalable Gaussian Processes with Billions of Inducing Inputs via 
  Tensor Train Decomposition}

\aistatsauthor{ Pavel A.~Izmailov \And Alexander V.~Novikov \And  
Dmitry A.~Kropotov }

\aistatsaddress{ 
  Lomonosov Moscow State\\ University, \\
  Cornell University\And
  National Research University\\ Higher School of Economics,\\
  Institute of Numerical Mathematics RAS\And
  Lomonosov Moscow State\\ University
}
]



\begin{abstract}
  We propose a method (TT-GP) for approximate inference in Gaussian Process (GP) 
  models. We build on previous scalable GP research including stochastic 
  variational inference based on inducing inputs, kernel interpolation, and 
  structure exploiting algebra.
  The key idea of our method is to use
  Tensor Train decomposition for variational parameters, which allows us to train
  GPs with billions of inducing inputs and achieve state-of-the-art results
  on several benchmarks. Further, our approach allows for training kernels based on
  deep neural networks without any modifications to the underlying GP model.
  A neural network learns a multidimensional embedding for the data, which is
  used by the GP to make the final prediction.
  without pretraining, through maximization of GP marginal likelihood.
  We show the efficiency of the
  proposed approach on several regression and classification benchmark datasets
  including MNIST, CIFAR-10, and Airline.

\end{abstract}

\section{Introduction}
  Gaussian processes (GPs) provide a prior over functions and allow finding complex
regularities in data. The ability of GPs to adjust the complexity of the model
to the size of the data makes them appealing to use for big datasets.
Unfortunately, standard methods for GP regression and classification scale as
$\bigO(n^3)$ with the number $n$ of training instances and can not be applied when $n$
exceeds several thousands.

Numerous approximate inference methods have been proposed in the literature. Many
of these methods are based on the concept of inducing inputs (\citet{candela2005},
\citet{snelson2006}, \citet{williams2000}). These methods build a smaller set
$Z$ of $m$ points that serve to approximate the true posterior of the process
and reduce the complexity to $\bigO(nm^2 + m^3)$. \citet{titsias2009} proposed
to consider the values $u$ of the Gaussian process at the inducing inputs
as latent variables and derived a variational inference procedure to approximate
the posterior distribution of these variables. \citet{hensman2013} and
\citet{hensman2015} extended this framework by using stochastic optimization to
scale up the method and generalized it to classification problems.

Inducing input methods allow to use Gaussian processes on datasets containing
millions of examples. However, these methods are still limited in the number
of inducing inputs $m$ they can use (usually up to $10^4$). Small number of 
inducing inputs limits the flexibility of the models that can be learned with 
these methods, and does not allow to learn expressive kernel functions 
(\citet{wilson2014}). \citet{wilson2015} proposed KISS-GP
framework, which exploits the Kronecker product structure in covariance matrices
for inducing inputs placed on a multidimensional grid in the feature space.
KISS-GP has complexity $\bigO(n + D m^{1 + 1/D})$, where $D$ is the dimensionality
of the feature space. Note however, that $m$ is the number of points in a
$D$-dimensional grid and grows exponentially with~$D$, which makes the method
impractical when the number of features $D$ is larger than $4$.

In this paper, we propose TT-GP method, that can use billions of inducing
inputs and is applicable to a much wider range of datasets compared to
KISS-GP.
We achieve this by combining kernel interpolation and Kronecker algebra of 
KISS-GP with a scalable variational inference procedure. We restrict the family of
variational distributions from \citet{hensman2013} to have parameters in
compact formats. Specifically, we use Kronecker product format for the
covariance matrix $\Sigma$ and Tensor Train format (\citet{oseledets2011}) for the
expectation $\mu$ of the variational distribution over the values $u$ of the
process at inducing inputs $Z$. 

\citet{nickson2015} showed that using Kronecker
format for $\Sigma$ does not substantially affect the predictive performance
of GP regression, while allowing for computational gains. The main contribution
of this paper is combining the Kronecker format for $\Sigma$ with TT-format
for $\mu$, which, together with efficient inference procedure, allows us to
efficiently train GP models with billions of inducing inputs.

Unlike KISS-GP the proposed method has linear complexity with respect to 
dimensionality $D$ of the feature space. It means that we can apply TT-GP
to datasets that are both large and high-dimensional. Note however, that
TT-GP is constructing a grid of inducing inputs in the feature space, and
tries to infer the values of the process in all points in the grid. 
High-dimensional real-world datasets are believed to lie on small-dimensional
manifolds in the feature space, and it is impractical to try to recover the
complex non-linear transformation that a Gaussian Process defines on the 
whole feature space. Thus, we use TT-GP on raw features for datasets with
dimensionality up to $10$. For feature spaces with higher dimensionality
we propose to use kernels based on parametric projections, which can be learned
from data.
 
\citet{wilson2016deep} and \citet{wilson2016stochastic}
demonstrated efficiency of Gaussian processes with kernels based on deep
neural networks. They used subsets of outputs of a DNN as
inputs for a Gaussian process. As the authors were using KISS-GP, they
were limited to using additive kernels, combining multiple low dimensional 
Gaussian processes. We found that DNN-based kernels are very efficient 
in combination with TT-GP. These kernels allows us to train TT-GP models
on high-dimensional datasets including computer vision tasks. Moreover,
unlike the existing deep kernel learning methods, TT-GP does not require
any changes in the GP model and allows deep kernels that produce embeddings
of dimensionality up to $10$.

\section {Background}
  \subsection{Gaussian Processes}
  A Gaussian process is a collection of random variables, any finite number of
  which have a joint normal distribution. A GP $f$ taking place in $\R^D$ is
  fully defined by its mean $m: \R^D \rightarrow \R$ and covariance
  $k: \R^D \times \R^D \rightarrow \R$ functions. For every $x_1, x_2, \ldots, x_n \in \R^D$
  \[
    f(x_1), f(x_2), \ldots, f(x_n) \sim \N(m, K),
  \]
  where $m = (m(x_1), m(x_2), \ldots, m(x_n))^T \in \R^n$, and
  $K\in \R^{n \times n}$ is the covariance matrix with $K_{ij}=k(x_i,x_j)$. Below we will use notation $K(A, B)$ for the matrix of pairwise
  values of covariance function $k$ on points from sets $A$ and $B$.

  Consider a regression problem. The dataset consists of
  $n$ objects $X = (x_1, \ldots, x_n)^T \in \R^{n \times D}$, and target values
  $y = (y_1, y_2, \ldots, y_n)^T \in \R^n$. We assume that the data is
  generated by a latent zero-mean Gaussian process $f$ plus independent Gaussian noise:
  \begin{equation}
    \begin{split}
      &p(y,f | X) = p(f | X)\prod_{i=1}^np(y_i | f_i),\\
      &p(f | X) = \N(f | 0, K(X,X)), \\
      &p(y_i | f_i) = \N(y_i | f_i, \nu^2 I),\\
    \end{split}
    \label{gp_model}
  \end{equation}
  where $f_i=f(x_i)$ is the value of the process at data point $x_i$ and $\nu^2$ is the noise variance.

  Assume that we want to predict the values of the process $f_*$ at a set of test
  points $X_*$. As the joint distribution of $y$ and $f_*$ is Gaussian, we can analytically
  compute the conditional distribution $p(f_* | y, X, X_*) = \N(f_*|\hat{m},\hat{K})$ 
  with tractable formulas for $\hat{m}$ and $\hat{K}$. The complexity of computing
  $\hat m$ and $\hat K$ is $\bigO(n^3)$ since it involves calculation of the 
  inverse of the covariance matrix $K(X, X)$.
  
  Covariance functions usually have a set of hyper-parameters $\theta$. For example, the RBF kernel
  \[
    k_{\mbox{\scriptsize RBF}} (x, x') = \sigma_f^2 \exp\left(- 0.5 \Vert x - x'\Vert^2 / l^2 \right)
  \]
  has two hyper-parameters $l$ and $\sigma_f$. In order to fit the model to the data,
  we can maximize the marginal likelihood of the process $p(y|X)$ with respect to these
  parameters. In case of GP regression this marginal likelihood is tractable and
  can be computed in $\bigO(n^3)$ operations.

  For two-class classification problem we use the same model~\eqref{gp_model} 
  with $p(y_i | f_i)=1/(1+\exp(-y_if_i))$, where $y_i\in\{-1,+1\}$. In this case
  both predictive distribution and marginal likelihood are intractable. For 
  detailed description of GP regression and classification see \citet{rasmussen2006}.

\subsection{Inducing Inputs}
\label{inducing_inputs}

  A number of approximate methods were developed to scale up Gaussian processes.
  \citet{hensman2013} proposed a variational lower bound that factorizes over
  observations for Gaussian process marginal likelihood. We rederive
  this bound here.

  Consider a set $Z \in \R^{m \times D}$ of $m$ inducing inputs in the feature 
  space and latent variables $u \in \R^m$ representing the values of
  the Gaussian process at these points. Consider the augmented model
  \[
    p(y, f, u) = p(y | f) p(f | u) p(u) = \prod_{i = 1}^ n p(y_i | f_i) p(f | u) p(u)
  \]
  with 
  \begin{equation}
  \label{priors}
  \begin{split}
    &p(f | u) = \N(f | \Knm \Kmm^{-1} u, \Knn - \Knm \Kmm^{-1} \Kmn),\\
    &p(u)= \N(u | 0, \Kmm),
  \end{split}
  \end{equation}
  where $\Knn = K(X, X), \Knm = K(X, Z), \Kmn = K(Z, X) = \Knm^T, \Kmm = K(Z, Z)$.

  The standard variational lower bound is given by
  \begin{equation}
  \label{standard_elbo}
  \begin{split}
    &\log p(y) \ge \E_{q(u, f)} \log \frac {p(y, f, u)} {q(u, f)} =\\
    & = \E_{q(f)} \log \prod_{i=1}^n p(y_i | f_i) - \KL(q(u, f) || p(u, f)),
  \end{split}
  \end{equation}
  where $q(u, f)$ is the variational distribution over latent variables.
  Consider the following family of variational distributions
  \begin{equation}
  \label{var_family}
    q(u, f) = p(f | u) \N(u | \mu, \Sigma),
  \end{equation}
  where $\mu \in \R^m$ and $\Sigma \in \R^{m \times m}$ are variational
  parameters. Then the marginal distribution over $f$ can be computed analytically
  \begin{equation}
  \begin{split}
    q(f)  = &\N\left(f | \Knm \Kmm^{-1} \mu,\right.\\
    &\left.\Knn + \Knm \Kmm^{-1} (\Sigma - \Kmm) \Kmm^{-1} \Kmn\right).
  \end{split}
  \end{equation}
  We can then rewrite (\ref{standard_elbo}) as
  \begin{equation}
  \label{main_elbo}
    \log p(y) \ge \sum_{i=1}^n  \E_{q(f_i)} \log p(y_i | f_i) - 
      \KL(q(u) || p(u)).
  \end{equation}
  Note, that the lower bound (\ref{main_elbo}) factorizes over observations and
  thus stochastic optimization can be applied to maximize this bound with respect
  to both kernel hyper-parameters $\theta$ and variational parameters $\mu$ and
  $\Sigma$, as well as other parameters of the model e.g. noise variance $\nu$.
  In case of regression we can rewrite (\ref{main_elbo}) in the closed
  form
  \begin{equation}
    \label{elbo_reg}
    \begin{split}
      \log & p(y) \ge 
      \sum_{i=1}^n  \bigg (\log \N (y_i | \ki^T \Kmm^{-1} \mu, \nu^2) - \\
        &- \frac 1 {2\nu^2} \tilde K_{ii} - 
        \left . \frac 1 {2\nu^2} \tr (\ki^T \Kmm^{-1} \Sigma \Kmm^{-1} \ki)
      \right )-
      \\
      - \frac 1 2 & \bigg(
      \log \frac {|\Kmm|} {|\Sigma|}  - m + \tr(\Kmm^{-1} \Sigma) +
        \mu^T \Kmm^{-1} \mu 
      \bigg),
    \end{split}
  \end{equation}
  where $\ki \in \R^m$ is the $i$-th column of $\Kmn$ matrix and $\tilde{K} = \Knn - \Knm \Kmm^{-1} \Kmn.$
  
  At prediction time we can use the variational distribution as a substitute for
  posterior
  \[
    p(f_* | y) = \int p(f_*| f, u) p(f, u |y) d f d u \approx
  \]
  \[
    \approx \int p(f_* | f, u) q(f, u) d f d u = \int p(f_* | u) q(u) du.
  \]

  The complexity of computing the bound (\ref{elbo_reg}) is $\bigO(n m^2 + m^3)$.
  \citet{hensman2015} proposes to use Gauss-Hermite quadratures to approximate the
  expectation term in (\ref{main_elbo}) for binary classification problem to
  obtain the same computational complexity $\bigO(nm^2 + m^3)$. This complexity
  allows to use Gaussian processes in tasks with millions of training samples,
  but these methods are limited to use small numbers of inducing inputs $m$,
  which hurts the predictive performance and doesn't allow to learn expressive
  kernel functions.

\subsection{KISS-GP}
\label{kiss_gp}

  \citet{saatci2012} noted that the covariance matrices computed at points on a
  multidimensional grid in the feature space can be represented as a Kronecker
  product if the kernel function factorizes over dimensions
  \begin{equation}
  \label{prod_kernel}
    k(x, x') = k_1(x^1, x'^1)\cdot k_2(x^2, x'^2)\cdot \ldots\cdot k_D(x^D, x'^D).
  \end{equation}
  Note, that many popular covariance functions, including RBF, belong to this class.
  Kronecker structure of covariance matrices allows to perform efficient inference
  for full Gaussian processes with inputs $X$ on a grid.

  \citet{wilson2015} proposed to set inducing inputs $Z$ on a grid:
  \[
    Z = Z^1 \times Z^2 \times \ldots \times Z^D,~~~~~Z^i \in \R^{m_i}~~~\forall i = 1, 2, \ldots, D.
  \]
  The number $m$ of inducing inputs is then given by $m = \prod_{i=1}^D m_{i}$.

  Let the covariance function satisfy (\ref{prod_kernel}). Then the covariance
  matrix $\Kmm$ can be represented as a Kronecker product over dimensions
  \[
    \Kmm = K_{m_1 m_1}^1 \otimes K^2_{m_2 m_2} \otimes \ldots \otimes
    K^D_{m_D m_D},
  \]
  where
  \[
    K^i_{m_i m_i} = K_i(Z_i, Z_i) \in \R^{m_i \times m_i}~~~\forall i = 1, 2, \ldots, D.
  \]
  Kronecker products allow efficient computation of matrix inverse and determinant:
  \begin{align*}
    &(A_1 \otimes A_2 \otimes \ldots \otimes A_D)^{-1} = A_1^{-1} \otimes A_2^{-1} \otimes \ldots \otimes A_D^{-1},\\
    &|A_1 \otimes A_2 \otimes \ldots \otimes A_D| = |A_1|^{c_1} \cdot |A_2|^{c_2} \cdot \ldots \cdot |A_D|^{c_D},
  \end{align*}
  where $A_i\in\R^{k_i \times k_i}$, $c_i = \prod_{j \ne i} k_j,\forall i = 1, 2, \ldots, D$.
  
  Another major idea of KISS-GP is to use interpolation to approximate $\Kmn$.
  Considering inducing inputs as interpolation points for the function
  $k(\cdot, z_i)$ we can write
  \begin{equation}
  \label{kernel_interpolation}
    \Kmn \approx \Kmm W,~~~\ki \approx \Kmm w_i,
  \end{equation}
  where $W \in \R^{m \times n}$ contains the coefficients of interpolation, and
  $w_i$ is it's $i$-th column. Authors of KISS-GP suggest using cubic
  convolutional interpolation (\citet{keys1981}), in which case the interpolation
  weights $w_i$ can be represented as a Kronecker product over dimensions
  \[
    w_i = w_i^1 \otimes w_i^2 \otimes \ldots \otimes w_i^D,~~~~~w_i \in \R^{m_i}~~~\forall i = 1, 2, \ldots, D.
  \]
  \citet{wilson2015} combine these ideas with SOR (\citet{silverman1985})
  in the KISS-GP method yielding $\bigO(n + D m^{1 + 1/D})$ computational
  complexity. This complexity allows to use KISS-GP with a large number (possibly
  greater than $n$) of inducing inputs. Note, however, that $m$ grows
  exponentially with the dimensionality $D$ of the feature space and the
  method becomes impractical when $D > 4$.

\subsection{Tensor Train Decomposition}
\label{tensor_train}

  Tensor Train (TT) decomposition, proposed in \citet{oseledets2011}, allows to
  efficiently store tensors (multidimensional arrays of data), large matrices, and
  vectors. For tensors, matrices and vectors in the TT-format linear algebra operations
  can be implemented efficiently. 

  Consider a $D$-dimensional tensor $\mathcal A \in \R^{n_1 \times n_2 \times \ldots \times n_D}$.
  $\mathcal{A}$ is said to be in the Tensor Train format if
  \begin{equation}
  \label{tt}
  \begin{split}
    \mathcal{A}(i_1, i_2, \ldots, i_D) =  G_1[i_1] \cdot G_2[i_2] \cdot \ldots \cdot G_D[i_D]&,\\
    i_d \in \{1, 2, \ldots, n_d\}~~\forall d&,
  \end{split}
  \end{equation}
  where $G_k[i_k] \in \R^{r_{k-1}\times r_{k}}\ \forall k, i_k,\ r_0 = r_{D} = 1$. Matrices $G_k$ are called TT-cores, and numbers $r_k$ are called TT-ranks of
  tensor $\mathcal{A}$.

  In order to represent a vector in TT-format, it is reshaped to a multidimensional
  tensor (possibly with zero padding) and then format (\ref{tt}) is used. We will
  use TT-format for the vector $\mu$ of expectations of the values $u$ of the
  Gaussian process in points $Z$ placed on a multidimensional grid. In this case,
  $\mu$ is naturally represented as a $D$-dimensional tensor.

  For matrices TT format is given by
  \[
    M(i_1, i_2, \ldots, i_d; j_1, j_2, \ldots, j_D) = 
  \]
  \[
    G_1 [i_1, j_1] \cdot
    G_2[i_2, j_2] \cdot \ldots \cdot G_D[i_D, j_D],
  \]
  where $G_k[i_k, j_k] \in \R^{r_{k-1}\times r_{k}}\ \forall k, i_k, j_k,\ r_0 = r_{D} = 1$.
  Note, that Kronecker product format is a special case of the TT-matrix with TT-ranks
  $r_1 = r_2 = \ldots = r_{D} = 1$.

  Let $u, v \in \R^{n_1 \cdot n_2 \cdot \ldots \cdot n_D}$ be vectors
  in TT-format with TT-ranks not greater than $r$. Let $A$ and $B$ be represented as a Kronecker product
  \[
    A = A_1 \otimes A_2 \otimes \ldots \otimes A_D,~~~A_k \in \R^{n_k \times n_k}~~\forall k,
  \]
  and the same for $B$. Let $n = \max_k n_k$. Then the computational complexity
  of computing the quadratic form $u^T A v$ is $\bigO(Dnr^3)$  and the 
  complexity of computing $\tr(AB)$ is
  $\bigO(Dn^2)$. We will need these two
  operations below. See \citet{oseledets2011} for a detailed description of
  TT format and efficient algorithms implementing linear algebraic operations
  with it.

\section{TT-GP}
  In the previous section we described several methods for GP regression and
classification. All these methods have different limitations: standard methods
are limited to small-scale datasets, KISS-GP requires small dimensionality of
the feature space, and other methods based on inducing inputs are limited to use
a small number $m$ of these points. In this section, we propose the TT-GP method
that can be used with big datasets and can incorporate billions of inducing
inputs. Additionally, TT-GP allows for training expressive deep kernels to work
with structured data (e.g. images).

\subsection{Variational Parameters Approximation}
  In section \ref{inducing_inputs} we derived the variational lower bound of
  \citet{hensman2013}. We will place the inducing inputs $Z$ on a
  multidimensional grid in the feature space and we will assume the
  covariance function satisfies (\ref{prod_kernel}). Let the number
  of inducing inputs in each dimension be $m_0$. Then the total number of inducing 
  inputs is $m = m_0^D$.
  As shown in Section \ref{kiss_gp},
  in this case $\Kmm$ matrix can be expressed as a Kronecker product over
  dimensions. Substituting the approximation (\ref{kernel_interpolation}) into
  the lower bound (\ref{elbo_reg}), we obtain
  \begin{equation}
  \label{kissgp_elbo}
    \begin{split}
      &\log p(y) \ge
        \\
        &\sum_{i=1}^n \bigg ( \log \N (y_i | w_i^T \mu, \nu^2) -
        \frac 1 {2\nu^2} \tilde K_{ii} - 
        \frac 1 {2\nu^2} \tr (w_i^T \Sigma w_i)
      \bigg )-
      \\
      &- \frac 1 2 \left(
        \log \frac {|\Kmm|} {|\Sigma|} - m + \tr(\Kmm^{-1} \Sigma) +
        \mu^T \Kmm^{-1} \mu
      \right),
    \end{split}
  \end{equation}
  where $\tilde K_{ii} = k(x_i, x_i) - w_i^T \Kmm w_i$.

  Note that $\Kmm^{-1}$ and $|\Kmm|$ can be computed with
  $\bigO(D m_0^3) = \bigO(D m^{3/D})$ operations due to the
  Kronecker product structure. Now the most computationally demanding terms
  are those containing variational parameters $\mu$ and $\Sigma$.

  Let us restrict the family of variational distributions~(\ref{var_family}). Let $\Sigma$ be
  a Kronecker product over dimensions, and $\mu$ be in the TT-format whith 
  TT-rank $r$ ($r$ is a hyper-parameter of our method). Then, according to 
  section \ref{tensor_train}, we can compute the lower bound~\eqref{kissgp_elbo}
  with
  $\bigO(nDm_0 r^2 + D m_0 r^3 + Dm_0^3) =
  \bigO(nD m^{1/D} r^2 + D m^{1/D} r^3 + D m^{3/D})$ complexity.

  The proposed TT-GP method has linear complexity with respect to dimensionality
  $D$ of the feature space, despite the exponential growth of the number of
  inducing inputs. Lower bound  (\ref{kissgp_elbo})
  can be maximized with respect to kernel hyper-parameters~$\theta$, TT-cores
  of $\mu$, and Kronecker multipliers of $\Sigma$. Note that stochastic optimization
  can be applied, as the bound (\ref{kissgp_elbo}) factorizes over data points.
  
  TT format was successfully applied for different machine learning tasks, 
  e.g. for compressing neural networks (\citet{novikov2015}) and
  estimating log-partition function in probabilistic graphical models 
  (\citet{novikov2014}). We explore the properties of approximating the 
  variational mean $\mu$ in TT format in section \ref{expect_approx}.

  \subsection{Classification}

  In this section we describe a generalization of the proposed method for
  multiclass classification. In this case the dataset consists of features
  $X = (x_1, x_2, \ldots, x_n)^T \in \R^{n \times D}$ and target values
  $y = (y_1, y_2, \ldots, y_n)^T \in \{1, 2, \ldots, C\}^n$, where $C$ is the
  number of classes.

  Consider $C$ Gaussian processes taking place in $\R^D$. Each process
  corresponds to it's own class. We will place $m = m_0^D$ inducing inputs $Z$ on a grid
  in the feature space, and they will be shared between all processes. Each
  process has it's own set of latent variables representing the values of
  the process at data points $f^c \in \R^n$, and inducing inputs $u^c \in \R^m$.
  We will use the following model
  \[
    p(y, f, u) = \prod_{i=1}^n p\left(y_i | f_i^{1, 2, \ldots, C}\right)
      \prod_{c=1}^C p\left(f^c | u^c\right) p(u^c),
  \]
  where $f_i^{1,2,\ldots,C}$ is the vector consisting of the values of all processes
  $1, 2, \ldots, C$ at data point $i$, $p(f^c | u^c)$ and $p(u^c)$ are defined 
  as in (\ref{priors}) and 
  \[
    p(y_i | f_i^{1,2, \ldots, C}) = 
    \frac {\exp(f_i^{y_i})} {\sum_{j=1}^C\exp(f_i^j)}.
  \]
  
  We will use variational distributions of the form
  \[
    q(f^1, f^2, \ldots, f^C, u^1, u^2, \ldots, u^C) =
  \]
  \[
    = q(f^1, u^1) \cdot q(f^2, u^2) \cdot \ldots \cdot q(f^C, u^C),
  \]
  where $q(f^c, u^c) = p(f^c | u^c) \N(u^c | \mu^c, \Sigma^c), c~=~1,~2,~\ldots,\\C$,
  all $\mu^c$ are represented in TT-format with TT-ranks not greater than $r$
  and all  $\Sigma^c$ are represented as Kronecker products over dimensions. 
  Similarly to (\ref{main_elbo}), we obtain
  \begin{equation}
  \label{elbo_multiclass}
  \begin{split}
    \log p(y) \ge & \sum_{i = 1}^n \E_{q(f_i^{1, 2, \ldots, C})} \log p(y_i | f_i^{1, 2, \ldots, C})
    \\
    & - \sum_{c = 1}^{C} \KL(q(u_c) || p(u_c))
  \end{split}
  \end{equation}
  The second term in (\ref{elbo_multiclass}) can be computed analytically as
  a sum of KL-divergences between normal distributions. The first term is
  intractable. In order to approximate the first term we will use a lower bound.
  We can rewrite
  \begin{equation}
  \label{expectation_term}
  \begin{split}
    &\E_{q\left(f_i^{1, 2, \ldots, C}\right)} \log p(y_i | f_i^{1, 2, \ldots, C}) =\\
    &\E_{q(f_i^{y_i})} f_i^{y_i} - \E_{q\left(f_i^{1, 2, \ldots, C}\right)}
    \log \biggl(\sum_{j=1}^C \exp (f_i^j) \biggr),
  \end{split}
  \end{equation}
  where $q(f_i^{1, 2\ldots, C}) = \N(f^1 | m_i^1, s_i^1)\cdot \N(f^2| m_2, s_i^2) \cdot \ldots \cdot \N(f^C | m_i^C, s_i^C)$.

  The first term in (\ref{expectation_term}) is obviously tractable, while the
  second term has to be approximated. \citet{bouchard2007} discusses several
  lower bounds for expectations of this type. Below we derive one of these bounds,
  which we use in TT-GP.

  Concavity of logarithm implies $\log\left(\sum_{j=1}^{C} \exp(f_i^j)\right) \le 
  \varphi \sum_{j=1}^C \exp(f_i^j) - \log \varphi - 1,\ \forall\varphi > 0$. 
  Taking expectation of both sides of the inequality and minimizing with respect
  to $\varphi$, we obtain
  \begin{equation}
  \label{bouchard_bound}
  \begin{split}
    \E_{q(f_i^{1, 2, \ldots, C})} \log\biggl(\sum_{j=1}^C \exp(f_i^j)\biggr) \le
    \\
    \log \biggl(\sum_{j=1}^C \exp\bigl(m_i^j + \frac 1 2 s_i^j\bigr) \biggr).
  \end{split}
  \end{equation}
  Substituting (\ref{bouchard_bound}) back into (\ref{elbo_multiclass}) we obtain
  a tractable lower bound for multiclass classification task, that can be
  maximized with respect to kernel hyper-parameters~$\theta^c$, TT-cores of
  $\mu^c$ and Kronecker factors of $\Sigma^c$. The complexity of the method
  is $C$ times higher, than in regression case.

\subsection{Deep kernels}

  \citet{wilson2016stochastic} and \citet{wilson2016deep} showed the efficiency
  of using expressive kernel functions based on deep neural networks with
  Gaussian processes on a variety of tasks. The proposed TT-GP method is
  naturally compatible with this idea.

  Consider a covariance function $k$ satisfying (\ref{prod_kernel}) and
  a neural network (or in fact any parametric transform) $net$. We can define a
  new kernel as follows
  \[
    k_{net}(x, x') = k(net(x), net(x')).
  \]
  We can train the neural network weights through maximization of GP marginal
  likelihood, the same way, as we normally train kernel hyper-parameters $\theta$.
  This way, the network learns a multidimensional embedding for the data, and
  GP is making the prediction working with this embedding.
  
  \citet{wilson2016stochastic} trained additive deep kernels combining 
  one-dimensional GPs on different outputs of a neural network. Training
  Gaussian processes on multiple outputs of a Neural network is impractical
  in their framework, because the complexity of the GP part of their model
  grows exponentially with the input dimensionality. 

  With methods of \citet{hensman2013} and \citet{hensman2015} training Gaussian
  processes on multiple outputs of a neural network also isn't straightforward. Indeed,
  with these methods we can only use up to $10^3$--$10^4$ inducing inputs. While
  with standard RBF kernels the positions of inputs of the GP are fixed and we can place 
  the inducing inputs near the data, with deep kernels the positions of the
  inputs of the GP (outputs of the DNN) change during training to match the positions of inducing inputs.
  It is thus not clear how to set the inducing inputs in the latent feature space, other
  than placing them on a multidimensional grid, which means that the complexity
  of such methods would grow exponentially with dimensionality.

  On the other hand, TT-GP allows us to train Gaussian processes on multiple DNN
  outputs because of it's ability to efficiently work with inducing inputs 
  placed on multidimensional grids.

\section{Experiments}
  In this section we first explore how well can we approximate variational
expectations in TT format with small ranks.
Then, we compare the proposed TT-GP
method with SVI-GP (\citet{hensman2013}) on regression tasks and KLSP-GP
(\citet{hensman2015}) on binary classification tasks using standard RBF kernel
functions. Then, we test the ability of our method to learn
expressive deep kernel functions and compare it with SV-DKL
(\citet{wilson2016stochastic}). For TT-GP we use our implementation available at
\url{https://github.com/izmailovpavel/TTGP}, which is based on the t3f library 
(\citet{novikov2018}). For SVI-GP and KLSP-GP we used the implementations provided in GPfLow
(\citet{GPflow2016}). 

\subsection{Expectation approximation}
\label{expect_approx}

In this section we provide a numerical justification of using Tensor Train
format for the mean $\mu$ of the variational distribution. We use the Powerplant
dataset from UCI. This dataset consists of $7654$ objects with $4$ features. We place 
$m_0 = 5$ inducing inputs per dimension and form a grid, which gives us a total of $m = 625$
inducing inputs. We train the standard SVI-GP method from GPflow library
(\citet{GPflow2016}) with free form representations for $\mu$ and $\Sigma$.
Then we try to approximate the learned $\mu$ vector with a TT-vector 
$\mu_{TT}$ with small TT-ranks. 

\begin{figure}[!h]
  \vspace{-.3cm}
  \begin{center}
      \begin{tabular}{cc}
          \hspace{-0.3 cm}\includegraphics[width=0.5\linewidth]{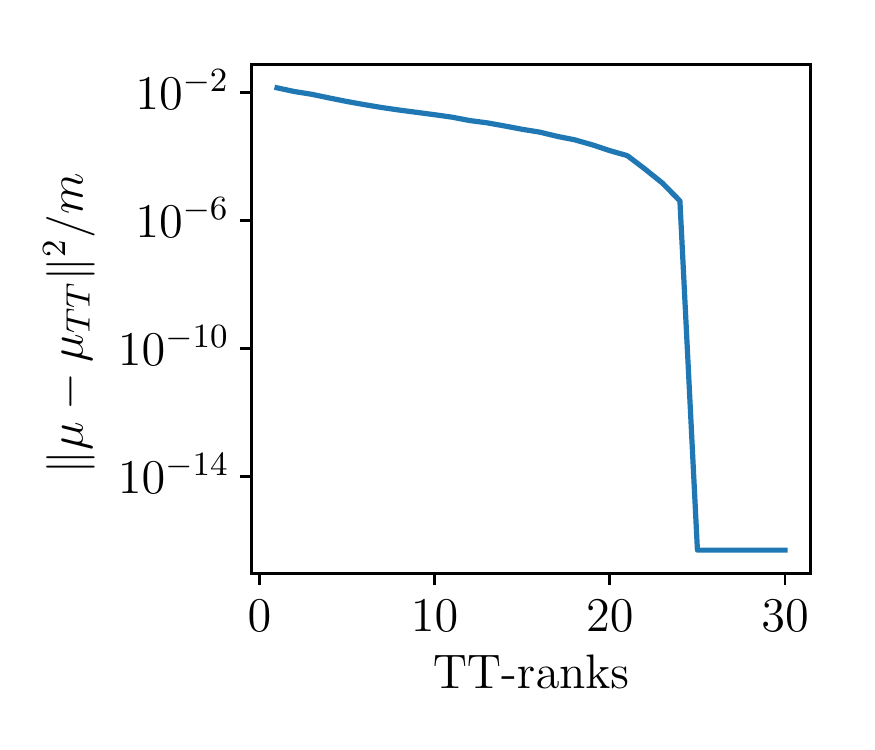} &
          \hspace{-0.3 cm}\includegraphics[width=0.5\linewidth]{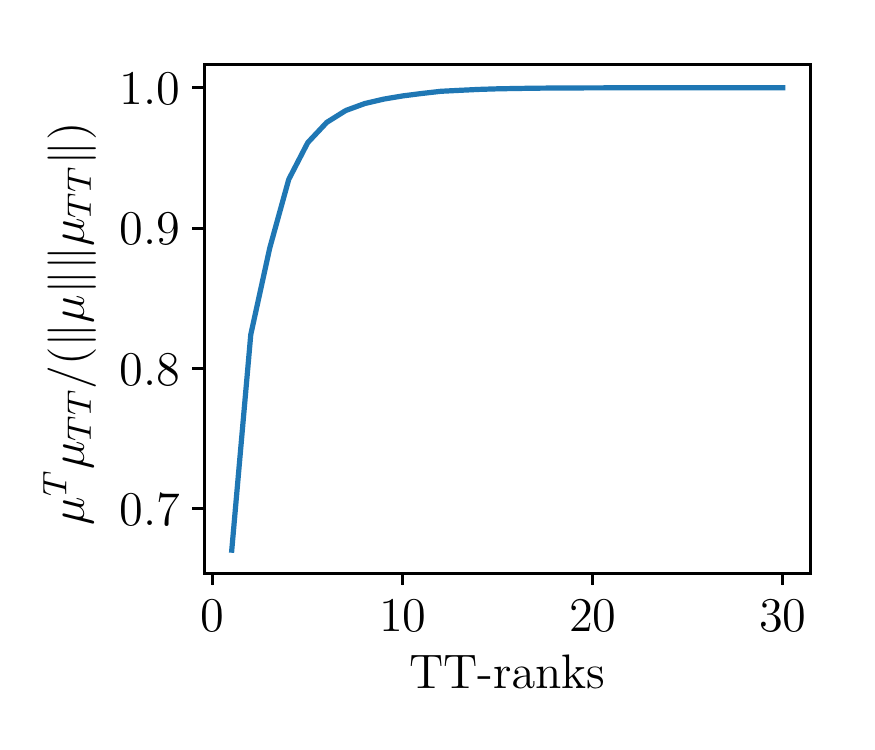} \\
          (a) MSE & 
          (b) Cosine similarity
      \end{tabular}
  \end{center}
  \caption{Approximation accuracy as a function of TT-rank.}
  \label{acc_vs_ranks}
\end{figure}

Figure \ref{acc_vs_ranks} shows the dependence between TT-ranks and 
approximation accuracy. For TT-rank greater than $25$ we can approximate the
true values of $\mu$ within machine precision. Note that for TT-rank $25$ the
amount of parameters in the TT representation already exceeds the number of
entries in the tensor $\mu$ that we are approximating. For moderate
TT-ranks an accurate approximation can still be achieved.

\begin{figure}[!h]
  \begin{center}
      \begin{tabular}{cc}
          \includegraphics[trim = 80 0 80 0, clip, height=0.4\linewidth]{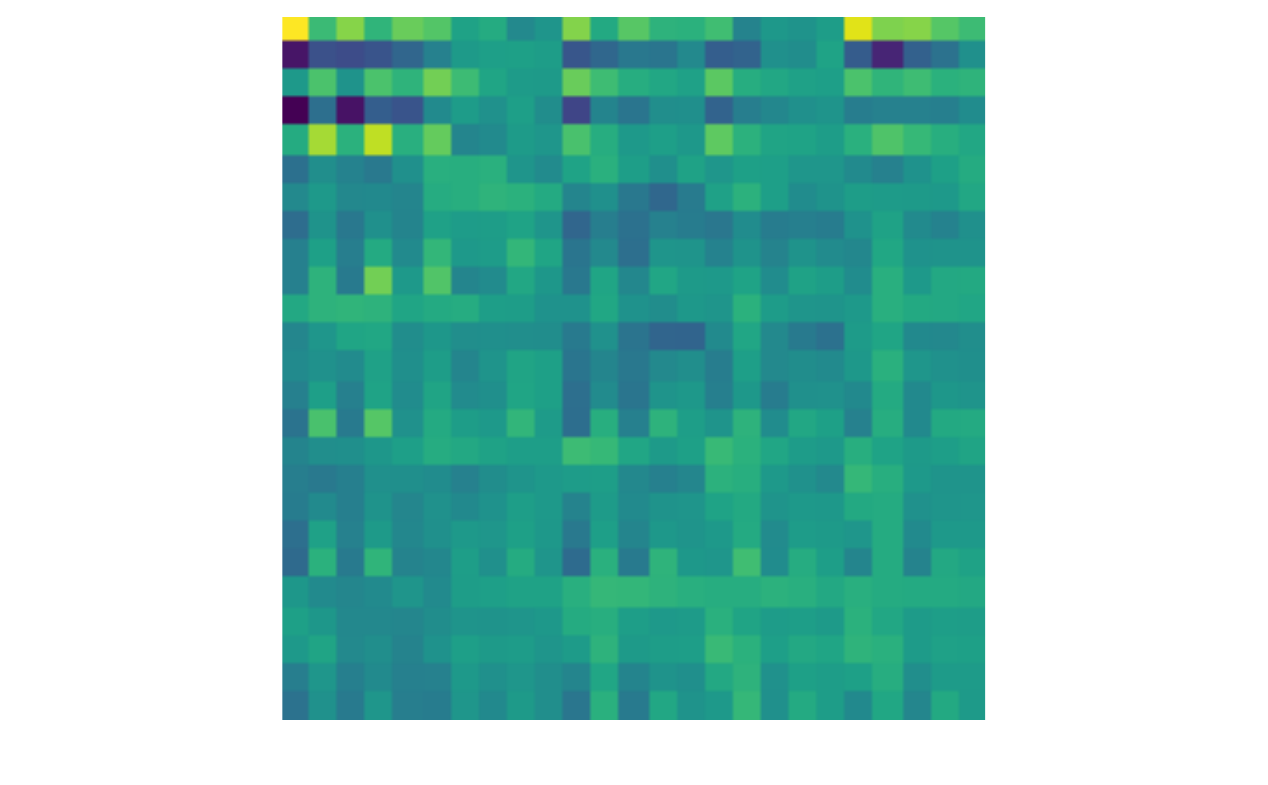} &
        \hspace{0.5cm}\includegraphics[trim = 80 0 0 0, clip, height=0.4\linewidth]{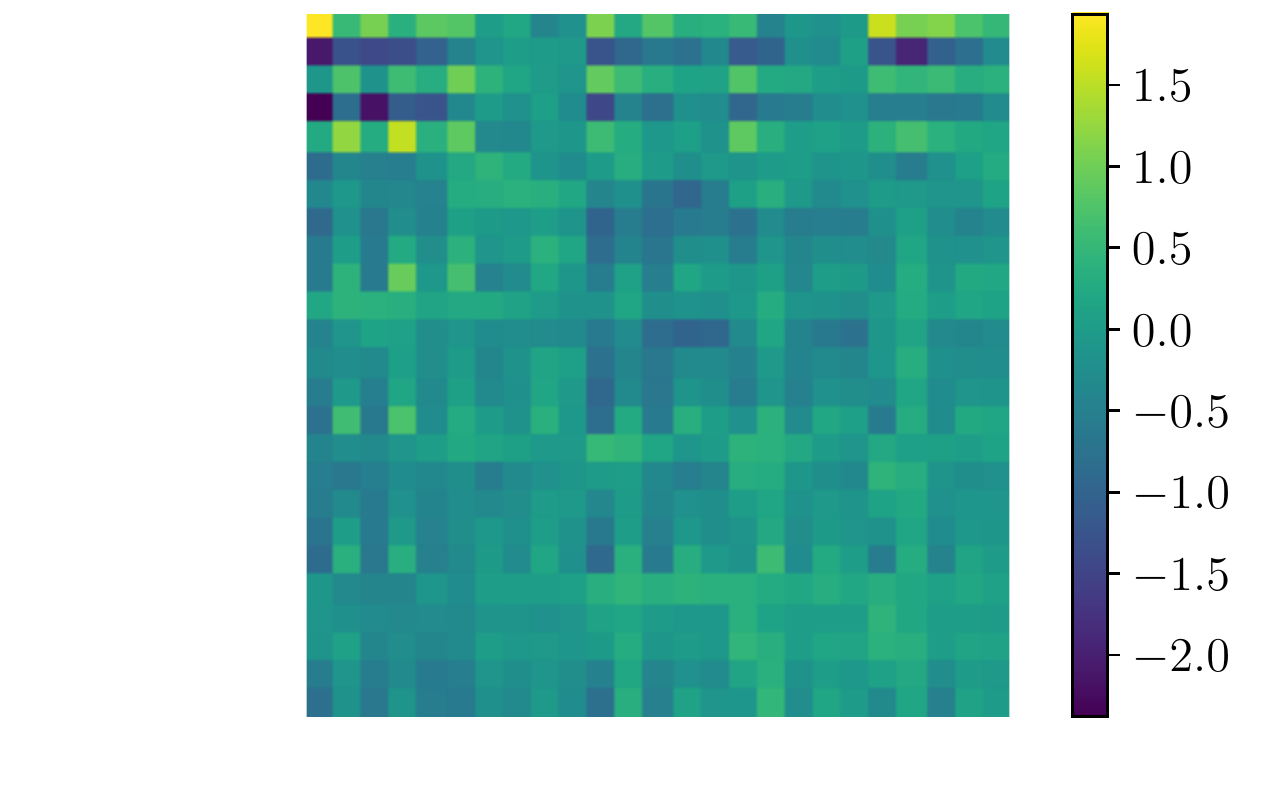} \\
          (a) $\mu$ &
          (b) $\mu_{TT}$, $r = 10$
      \end{tabular}
  \end{center}
  \vspace{-0.4cm}
  \caption{True variational mean and TT-approximation. Here we reshape the 
    $625$-dimensional $\mu$ and $\mu_{TT}$ vectors
    to $25 \times 25$ matrices for visualization.}
  \label{true_and_tt}
  \vspace{-0.2cm}
\end{figure}

Figure \ref{true_and_tt} shows the true variational mean $\mu$ and it's 
approximation for TT-rank $10$.  We can see that $\mu_{TT}$ captures the 
structure of the true variational mean.

\begin{table*}[!t]
  \caption[]{Experimental results for standard RBF kernels. In the table acc. stands for $r^2$ for regression and accuracy
          for classification tasks. $n$ is the size of the
          training set, $D$ is the dimensionality of the feature space,
          $m$ is the number of inducing inputs, $r$ is TT-ranks of $\mu$ for TT-GP; $t$ is the time per one pass over
          the data (epoch) in seconds; where provided, $d$ is the dimensionality of linear embedding.\\
          $^*$ for KLSP-GP on Airline we provide results from the original
          paper where the accuracy is given as a plot, and detailed
          information about experiment setup and exact results is not available.
          }
  \label{se_results}
  \centering
  \begin{tabular}{lll l cll l cllll}
    \toprule
    \multicolumn{3}{c}{Dataset} && \multicolumn{3}{c}{SVI-GP / KLSP-GP} && \multicolumn{5}{c}{TT-GP} \\
    \cmidrule{1-3}
    \cmidrule{5-7}
    \cmidrule{9-13}

    Name & $n$ & $D$ &&
    acc. & $m$ & $t$ (s) &&
    acc. & $m$ & $r$ & $d$ & $t$ (s)\\
    \midrule

    Powerplant & $7654$ & $4$ &&
    $0.94$ & $200$ & $10$ &&
    $0.95$ & $35^4$ & $30$ & - & $5$ \\

    Protein & $36584$ & $9$ &&
    $0.50$ & $200$ & $45$ &&
    $0.56$ & $30^9$ & $25$ & - & $40$ \\

    YearPred & $463K$ & $90$ &&
    $0.30$ & $1000$ & $597$ &&
    $0.32$ & $10^6$ & $10$ & $6$ & $105$ \\

    \midrule
    Airline & $6M$ & $8$ &&
    $0.665^*$ & - & - &&
    $0.694$ & $20^8$ & $15$ & - & $5200$ \\

    svmguide1 & 3089 & 4 &&
    $0.967$ & $200$ & $4$ &&
    $0.969$ & $20^4$ & $15$ & - & $1$\\

    EEG & 11984 & 14 &&
    $0.915$ & $1000$ & $18$ &&
    $0.908$ & $12^{10}$ & $15$ & $10$ & $10$\\

    covtype bin & 465K & 54 &&
    $0.817$ & $1000$ & $320$ &&
    $0.852$ & $10^6$ & $10$ & $6$ & $172$\\
    \bottomrule
  \end{tabular}
\end{table*}

\subsection{Standard RBF Kernels}

For testing our method with standard RBF covariance functions we used a range of
classification and regression tasks from UCI and LIBSVM archives and the
Airline dataset, that is popular for testing scalable GP models
(\citet{hensman2013}, \citet{hensman2015}, \citet{wilson2016stochastic},
\citet{cutajar2016}).

For Airline dataset we provide results reported in
the original paper (\citet{hensman2015}). For our experiments, we use a cluster of Intel Xeon E5-2698B v3 CPUs having $16$ cores and $230$ GB
of RAM.

For YearPred, EEG and covtype datasets we used a $d$-dimensional linear embedding inside the RBF kernel for TT-GP, as the number $D$ of features makes it
impractical to set inducing inputs on a grid in a $D$-dimensional space in this case.

Table \ref{se_results} shows the results on different regression and
classification tasks. We can see, that TT-GP is able to achieve better
predictive quality on all datasets except EEG. We also note that the
method is able to achieve good predictive performance with linear
embedding, which makes it practical for a wide range of datasets.

\subsection{Deep Kernels}

  \subsubsection{Representation learning}
\begin{figure}[!t]
  \begin{center}
      \begin{tabular}{c}
          \hspace{-.9cm}\includegraphics[height=0.45\linewidth]{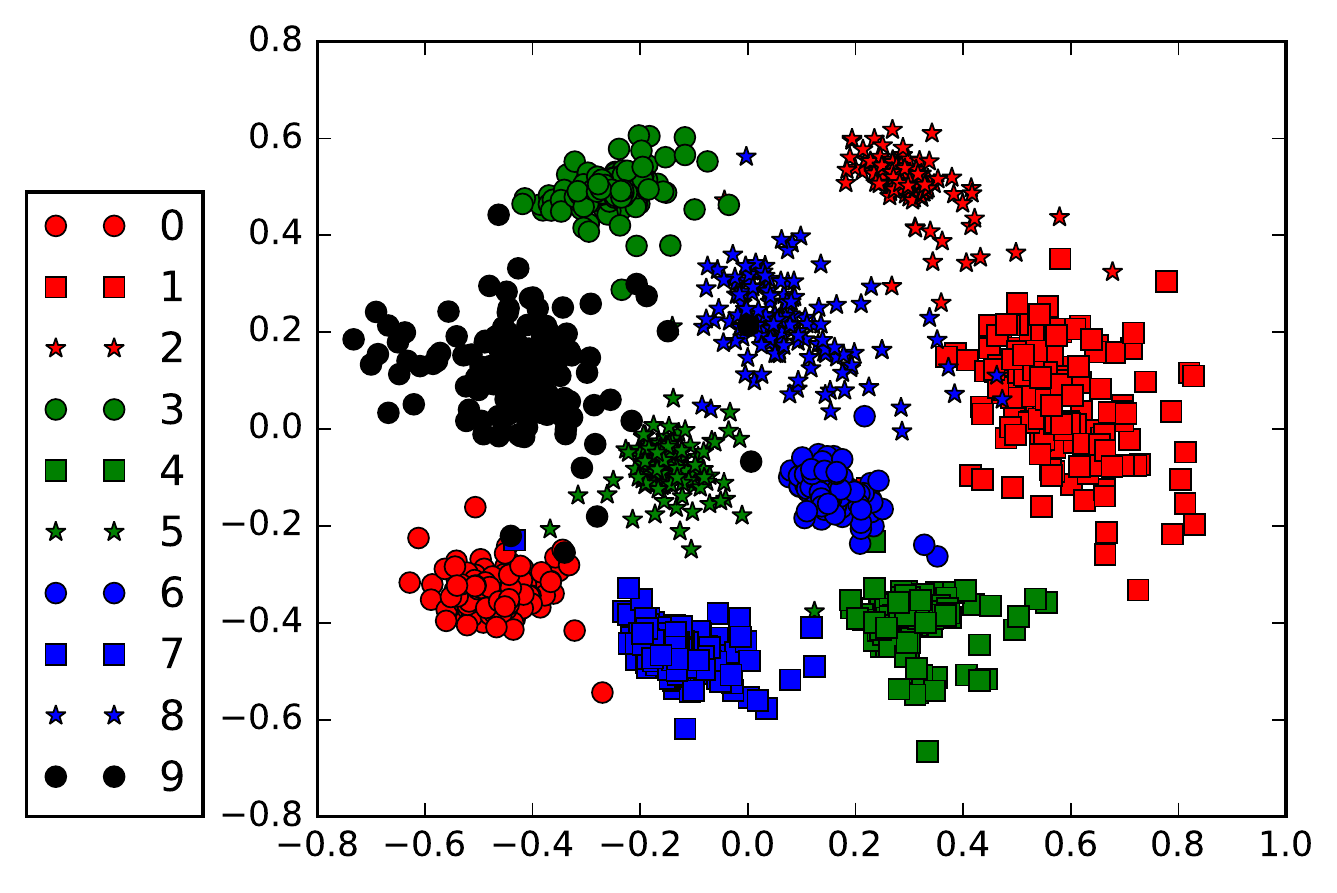} \\
          (a) DNN with TT-GP \\
          \includegraphics[height=0.45\linewidth]{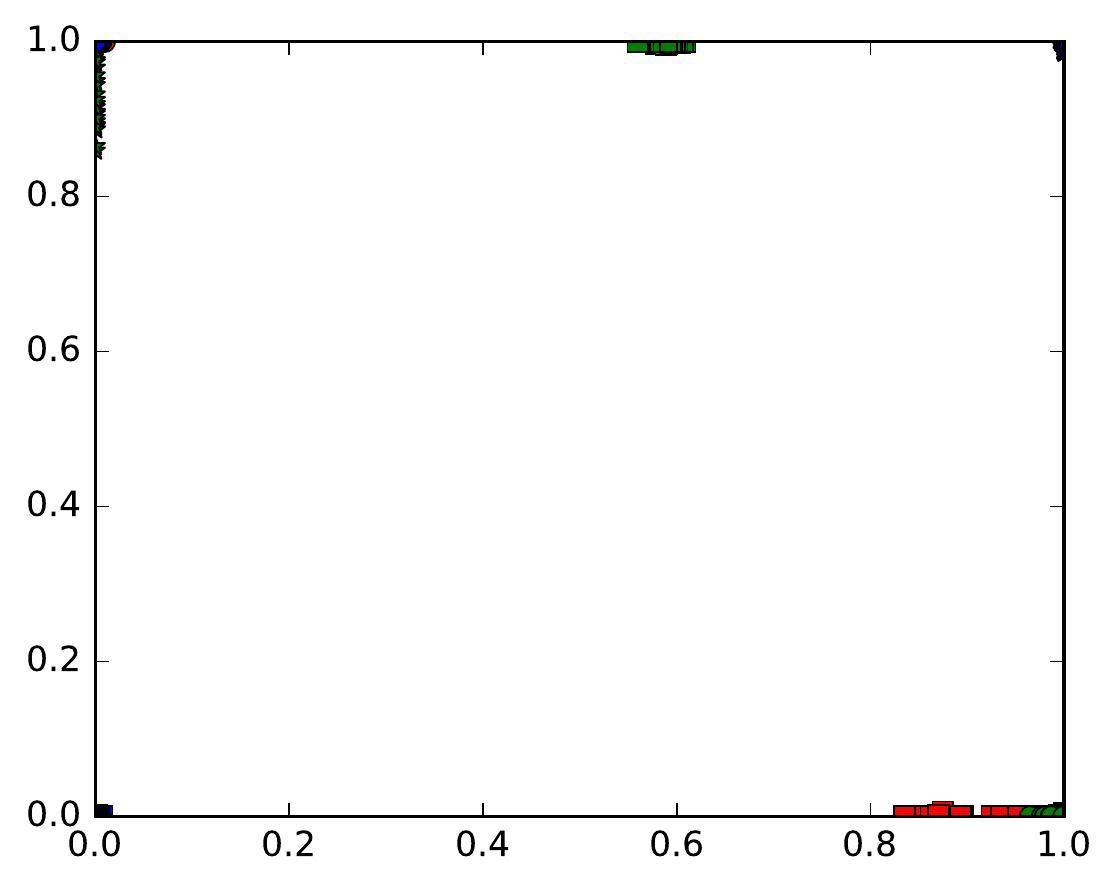}\\
          (b) Plain DNN\\
      \end{tabular}
  \end{center}
  \vspace{-0.4cm}
  \caption{Learned representation for Digits dataset.}
  \vspace{-0.4cm}
  \label{digits_embedding}
\end{figure}

  We first explore the representation our model learns for data on the small
  Digits\footnote{\url{http://scikit-learn.org/stable/auto_examples/datasets/plot_digits_last_image.html}}
  dataset containing $n = 1797$ $8 \times 8$ images of handwritten digits. We
  used a TT-GP with a kernel based on a small fully-connected neural network
  with two hidden layers with $50$ neurons each and $d = 2$ neurons in the output
  layer to obtain a $2$-dimensional embedding. We trained the model to classify
  the digits to $10$ classes corresponding to different digits.
  Fig. \ref{digits_embedding} (a) shows the learned embedding. We also trained the same
  network standalone, adding another layer with $10$ outputs and softmax
  activations. The embedding for this network is shown in fig. \ref{digits_embedding},b.

  \begin{table*}[!t]
    \caption{DNN architecture used in experiments with deep kernels. Here F($h$) means a fully-connected layer with $h$ neurons; C($h{\times}w$, $f$) means a
    convolutional layer with $f$ $h{\times}w$ filters; P($h{\times}w$) means max-pooling with $h{\times}w$ kernel; ReLU stands for rectified linear unit
    and BN means batch normalization (\citet{ioffe2015}).}
    \label{deep_architecture}
    \centering
    \begin{tabular}{ll}
      \toprule
      Dataset & Architecture \\
      \midrule

      Airline & F($1000$)-ReLU-F($1000$)-ReLU-F($500$)-ReLU-F($50$)-ReLU-F($2$)\\

      CIFAR-10 & C($3{\times}3$, 128)-BN-ReLU-C($3{\times}3$, 128)-BN-ReLU-P($3{\times}3$)-C($3{\times}3$, 256)-BN-ReLU-\\
       & C($3{\times}3$, 256)-BN-ReLU-P($3{\times}3$)-C($3{\times}3$, 256)-BN-ReLU-\\
       & C($3{\times}3$, 256)-BN-ReLU-P($3{\times}3$)-F($1536$)-BN-ReLU-F($512$)-BN-ReLU-F($9$)\\

      MNIST & C($5{\times}5$, $32$)-ReLU-P($2{\times}2$)-C($5{\times}5$, $64$)-ReLU-P($2{\times}2$)-F($1024$)-ReLU-F($4$)\\
      \bottomrule
    \end{tabular}
  \end{table*}
  
  \begin{table*}[t]
    \caption{Results of experiments with deep kernels. Here acc. is classification
            accuracy; $C$ is the number of classes; $d$ is the dimensionality
            of embedding learned by the model; $t$ is the time per one pass over
            data (epoch) in seconds.}
    \label{deep_results}
    \centering
    \begin{tabular}{llll ll llll lll}
      \toprule
      \multicolumn{4}{c}{Dataset}  && SV-DKL &&
      \multicolumn{2}{c}{DNN} &&
      \multicolumn{3}{c}{TT-GP}\\

      \cmidrule{1-4}
      \cmidrule{6-6}
      \cmidrule{8-9}
      \cmidrule{11-13}

      Name & $n$ & $D$ & $C$ &&
      acc. && acc. & $t$ (s) &&
      acc. & $d$ & $t$ (s)
      \\
      \midrule


      Airline & $6M$ & $8$ & $2$ &&
      $0.781$ && $0.780$ & $1055$ &&
      $0.788 \pm 0.002$ & $2$ & $1375$\\

      CIFAR-10 & $50K$ & $32{\times}32{\times}3$ & $10$ &&
      $0.770$ && $0.915$ & $166$ &&
      $0.908 \pm 0.003$ & $9$ & $220$\\

      MNIST & $60K$ & $28{\times}28$ & $10$ &&
      $0.992$ && $0.993$ & $23$ &&
      $0.9936 \pm 0.0004$ & $10$ & $64$\\
      \bottomrule
    \end{tabular}
  \end{table*}

  We can see that the stand-alone DNN with linear classifiers is unable
  to learn a good $2$-dimensional embedding.
  On the other hand, using a flexible GP classifier that is capable of learning
  non-linear transormations, our
  model groups objects of the same class into compact regions.

  \subsubsection{Classification tasks}
  To test our model with deep kernels we used Airline,
  CIFAR-10 (\citet{krizhevsky2009}) and
  MNIST (\citet{lecun1998}) datasets. The corresponding DNN architectures are shown
  in Table~\ref{deep_architecture}. For CIFAR-10 dataset we also use standard 
  data augmentation techniques with random cropping of $24 \times 24$
  parts of the image, horizontal flipping, randomly adjusting brightness and contrast. In all experiments we also add a BN without trainable mean and
  variance after the DNN output layer to project the outputs into the region
  where inducing inputs are placed. We use $m_0 = 10$ inducing inputs
  per dimension placed on a regular grid from $-1$ to $1$ and set TT-ranks of 
  $\mu$ to $r = 10$ for all three datasets. For experiments with convolutional neural networks, we
  used Nvidia Tesla K80 GPU to train the model.

  Table~\ref{deep_results} shows the results of the experiments for our TT-GP
  with DNN kernel and SV-DKL. Note, that the comparison
  is not absolutely fair on CIFAR-10 and MNIST datasets, as we didn't use
  the same exact architecture and preprocessing as \citet{wilson2016stochastic}
  because we couldn't find the exact specifications of these models.
  On Airline dataset we used the same exact architecture and preprocessing as
  SV-DKL and TT-GP achieves a higher accuracy on this dataset.

  We also provide results of stand-alone DNNs for comparison. We used the
  same networks that were used in TT-GP kernels with the last linear layers replaced
  by layers with $C$ outputs and softmax activations. Overall, we can see, that
  our model is able to achieve good predictive performance,
  improving the results of standalone DNN on Airline and MNIST.

  We train all the models from random initialization without pretraining. We also
  tried using pretrained DNNs as initialization for the kernel of our TT-GP model,
  which sometimes leads to faster convergence, but does not improve the final
  accuracy.

\section{Discussion}
  
We proposed TT-GP method for scalable inference in Gaussian process models
for regression and classification.  The proposed method is capable of using
billions of inducing inputs, which is impossible for existing methods. This allows us to improve the
performance over state-of-the-art both with standard and deep kernels
on several benchmark datasets.
Further, we believe that our model provides a more natural way of learning deep
kernel functions than the existing approaches since it doesn't require any
specific modifications of the GP model and allows working with high-dimensional
DNN embeddings.
 
Our preliminary experiments showed that TT-GP is inferior in terms of
uncertainty quantification compared to existing methods. We suspect that the
reason for this is restricting Kronecker structure for the covariance matrix $\Sigma$. 
We hope to alleviate this limitation by using Tensor Train format for $\Sigma$ 
and corresponding approximations to it's determinant.

As a promising direction for future work we consider training TT-GP
with deep kernels incrementally, using the variational approximation
of posterior distribution as a prior for new data. We also find it interesting
to try using the low-dimensional embeddings learned by our model for transfer 
learning. Finally, we are interested in using the proposed method for structured prediction,
where TT-GP could scale up GPstruct approaches (\citet{bratieres2015}) and allow using
deep kernels. 

\subsubsection*{Acknowledgements}

Alexander Novikov was supported by the Russian Science Foundation grant 17-11-01027.
Dmitry Kropotov was supported by Samsung Research, Samsung Electronics.

\bibliography{bib/biblio}
\bibliographystyle{plainnat}

\end{document}